%% file: main.tex
\documentclass[conference, 10pt]{IEEEtran}
\usepackage[left=54pt, right=54pt, bottom=54pt, top=54pt]{geometry}
\IEEEoverridecommandlockouts
\usepackage{cite}
\usepackage{amsmath,amssymb,amsfonts}
\def\IEEEtitletopspace{18pt}
\usepackage{bbold}
\usepackage{graphicx}
\usepackage{textcomp}
\usepackage{mathtools}
\def\BibTeX{{\rm B\kern-.05em{\sc i\kern-.025em b}\kern-.08em
    T\kern-.1667em\lower.7ex\hbox{E}\kern-.125emX}}
    
\usepackage[pagebackref=false,breaklinks=true,colorlinks=false,citecolor=black,bookmarks=false]{hyperref}
\include{nomenclature}
\newcommand{\ours}{\mbox{GP-net}}

\begin{document}

\title{\ours: Flexible Viewpoint Grasp Proposal\\
\thanks{This publication has emanated from research conducted with the financial support of Science Foundation Ireland under Grant numbers 18/CRT/6049 and 16/RI/3399. For the purpose of Open Access, the author has applied a CC BY public copyright licence to any Author Accepted Manuscript version arising from this submission. The opinions, findings and conclusions or recommendations expressed in this material are those of the author(s) and do not necessarily reflect the views of the Science Foundation Ireland.}}

\author{\IEEEauthorblockN{Anna Konrad}
\IEEEauthorblockA{\textit{Hamilton Institute} \\
\textit{Maynooth University}\\
anna.konrad.2020@mumail.ie}
\and
\IEEEauthorblockN{John McDonald}
\IEEEauthorblockA{\textit{Department of Computer Science} \\
\textit{Maynooth University}\\
john.mcdonald@mu.ie}
\and
\IEEEauthorblockN{Rudi Villing}
\IEEEauthorblockA{\textit{Department of Electronic Engineering} \\
\textit{Maynooth University}\\
rudi.villing@mu.ie}
}

\maketitle

\begin{abstract}
We present the Grasp Proposal Network (\ours), a Convolutional Neural Network model which can generate \mbox{6-DoF} grasps from flexible viewpoints, e.g. as experienced by mobile manipulators. To train \ours, we synthetically generate a dataset containing depth-images and ground-truth grasp information. In real-world experiments, we use the EGAD evaluation benchmark to evaluate \ours\ against two commonly used algorithms, the Volumetric Grasping Network (VGN) and the Grasp Pose Detection package (GPD), on a PAL TIAGo mobile manipulator. In contrast to the state-of-the-art methods in robotic grasping, \ours\ can be used for grasping objects from flexible, unknown viewpoints without the need to define the workspace and achieves a grasp success of $54.4\%$ compared to $51.6\%$ for VGN and $44.2\%$ for GPD. We provide a ROS package along with our code and pre-trained models at \url{https://aucoroboticsmu.github.io/GP-net/}. 
\end{abstract}


\section{Introduction}\label{sec:intro}

Manipulation is an ongoing and challenging problem in robotics due to its complexity and variability. Grasping objects is largely solved in industrial settings with known objects and fixed, foreknown poses of the object and robot. However, more complex, dynamic and unstructured environments are still an active field of research. Before the era of machine learning, analytical approaches were widely tested in simulation but had limited applicability in real-world scenarios due to noisy or partial data~\cite{bohg2014}. With the rise of computational power, researchers have switched to data-driven approaches, which are more robust to real-world conditions. To simplify the problem, these solutions initially focused on fixed overhead cameras and a reduced grasp space with 4 Degrees-of-Freedom (DoF) grasps~\cite{mahler2017dexnet, zhang2020randomforest, satish2019policy, morrison2019, redmon2015, kumra2020}, i.e. top-grasps which can vary their pose in 3D position and one rotational axis.

However, when grasping outside of a lab environment, the restriction to top-grasps limits the applicability of the algorithms, while \mbox{6-DoF} grasps can enhance reachability for the robots. Furthermore, if the camera position is not fixed, camera pose changes must be handled. Finally, methods must be robust to sensor noise in real-world environments. Efforts are being made towards solutions for many of these problems, for example with \mbox{6-DoF} grasp proposal algorithms~\cite{breyer2020vgn, sundermeyer2021contactgraspnet, lundell2020scenecompletion, pas2017pointclouds, berscheid20216dof}, closed-loop grasping~\cite{morrison2019, levine2018, viereck2017}, or grasping transparent objects~\cite{IchnowskiAvigal2021DexNeRF}. 

\begin{figure}[t]
      \centering
      \includegraphics[width=0.45\textwidth]{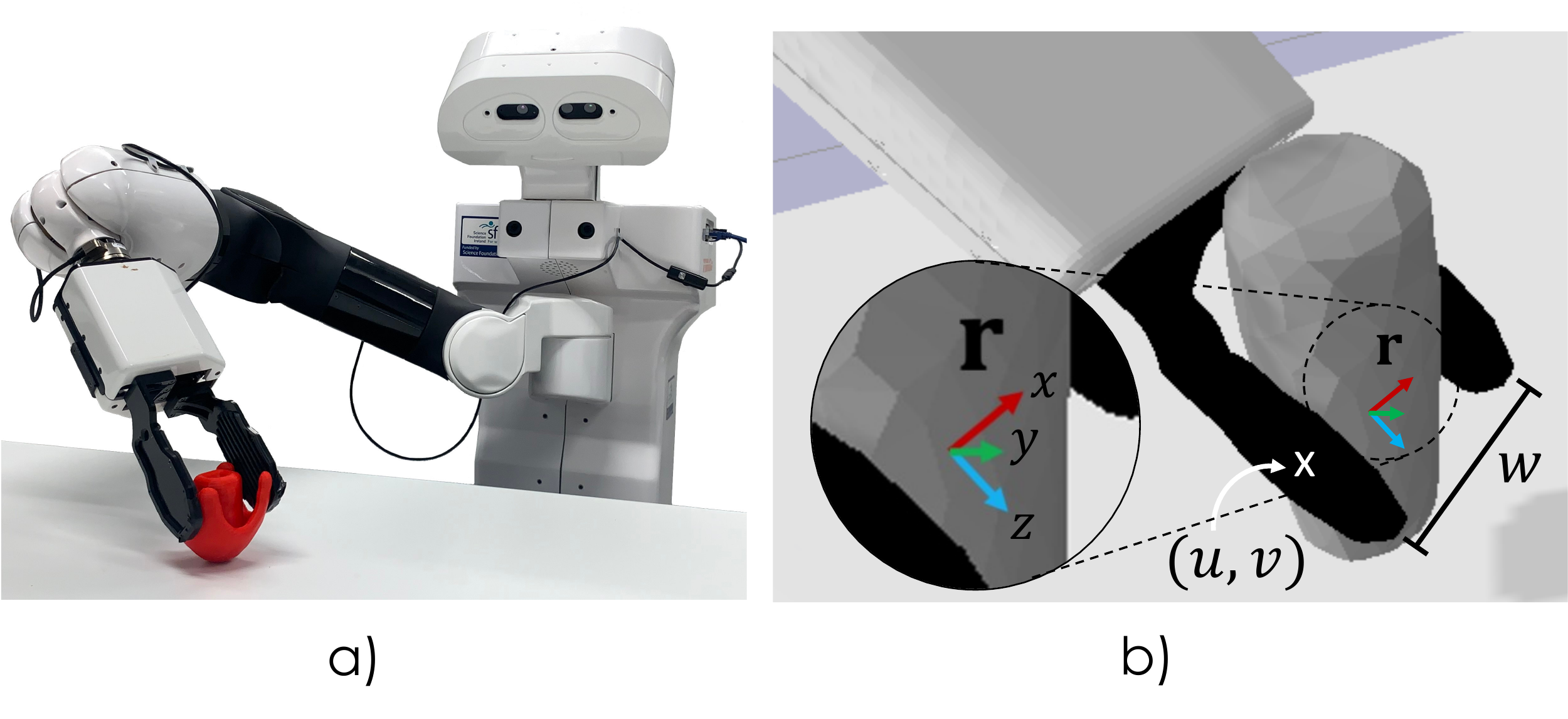}
      \caption{a) A {PAL} TIAGo robot grasping an object. b) Visualisation of the contact-based grasp representation with visible grasp contact coordinates in the image $(u, v)$, grasp orientation $\mathbf{r}$ and grasp width $w$.}
      \label{fig:grasp}
\end{figure}

Applying existing algorithms to mobile manipulators (robots that have a mobile base) often proves difficult. Both the robot and its camera can move, resulting in the need to operate from a variety of viewpoints with respect to the environment and objects to be grasped and manipulated. Furthermore, the environment tends to be dynamic and uncertain and, in particular, the robot's workspace is not known in advance as it is for fixed manipulators. While there exist algorithms that can be used to propose grasps on mobile robots, they typically require some sort of workspace definition to  improve the grasp proposals~\cite{pas2017pointclouds} or to run the algorithm itself~\cite{breyer2020vgn}. Other solutions require wrist-mounted depth cameras for visuomotor control~\cite{viereck2017} or high-end GPUs to run inference~\cite{sundermeyer2021contactgraspnet}.

We present the grasp proposal network (\ours), a model that can predict \mbox{6-DoF} grasps on single objects. Unlike other approaches, \ours\ operates directly on the input depth image without requiring a prior workspace definition and can thus handle flexible and unknown camera viewpoints. \ours\ can be used to grasp novel objects from planar surfaces like tables or furniture units with mobile manipulators. In real-world experiments using the EGAD evaluation benchmark~\cite{morrison2020egad} with a PAL TIAGo manipulator, \ours\ achieves a grasp success of $54.4\%$, performing slightly better than the Volumetric Grasping Network (VGN) with $51.6\%$ and substantially outperforming the Grasp Pose Detection package (GPD) with $44.2\%$. 

The contributions of this work can be summarised as follows:
\begin{itemize}
    \item \ours: a model to predict \mbox{6-DoF} grasps for a parallel jaw gripper from flexible viewpoints without specifying the workspace or running table segmentation to filter grasp poses.
    \item A ROS package to run \ours\ to produce grasp proposals from a depth camera.
    \item A dataset to train \ours\ or alternative network architectures. We make our code available so it can be used to train \ours\ to adapt to different robots or grippers.
\end{itemize}

\section{Background and Related Work}\label{sec:background}

Grasping objects in real-world settings involves a lot of variation and uncertainty, including unknown object shapes, unknown poses of the robot and objects, and noisy sensor inputs. To solve this high-dimensional problem, early analytical approaches constrained and simplified the problem by making assumptions about the contacts, friction, and the geometric and physical model of the objects. Consequently, these approaches were usually only tested in simulation with accurate knowledge about the object, robot and environment and without the need to contend with noisy sensor measurements~\cite{bohg2014}.

With the surge of data-driven approaches, analytical methods have been surpassed by machine learning algorithms for grasp synthesis~\cite{bohg2014}. Initially, these methods typically used {4-DoF} top-grasps, where the gripper approaches the object top-down and can only rotate around the approach axis. Furthermore, the initial techniques were usually discriminative~\cite{kleeberger2020survey}, where grasp configurations are sampled and subsequently ranked according to a quality metric~\cite{mahler2017dexnet, lenz2015, zhang2020randomforest, asif2018ensemblenet}. While discriminative methods showed good grasping performance, algorithms directly proposing suitable grasp configurations in a generative manner can improve inference time significantly and enable grasping in real-time~\cite{redmon2015, kumra2020, morrison2019, satish2019policy, breyer2020vgn}.

To make robotic grasping usable in less constrained environments, the algorithms must be able to propose \mbox{6-DoF} grasp poses while coping with varying camera viewpoints and object poses. These requirements increase the search space for possible grasps and, therefore, the problem's difficulty significantly. While modern algorithms can find robust grasp poses in this increased search space, they often suffer from issues such as high latency~\cite{varley2017shapecompletion}, high computational requirements~\cite{sundermeyer2021contactgraspnet}, or limitations to the sampling workspace~\cite{breyer2020vgn}. 

Generally, comparing the performance of different methods proves difficult due to the different robots, grippers and objects being used. Usually, algorithms are tested on household objects without the opportunity to reproduce the experiments or compare performance~\cite{breyer2020vgn, sundermeyer2021contactgraspnet, mahler2017dexnet, varley2017shapecompletion, lundell2020scenecompletion, jiang2021synergies, chen2021transsc, pas2017pointclouds}. More recently, benchmarks like EGAD~\cite{morrison2020egad} using 3D-printed objects have been proposed to generalise the testing of grasping algorithms and improve the comparability and reproducibility of experiments. For this reason, we use EGAD objects in our real-world experiments (see Section~\ref{subsec:results}).

One of the first methods that could be directly applied to find \mbox{6-DoF} grasp poses based on point-clouds is the Grasp Pose Detection (GPD) package~\cite{pas2017pointclouds}. It identifies a region of interest, samples grasp proposals, encodes them in a multi-channel image, and finally uses a Convolutional Neural Network to predict the grasp quality. While it is widely used due to its availability in a {ROS} package, we find that it mainly suggests grasps on the table plane if the table segmentation fails. Further, the discriminative sampling nature of the algorithm leads to high latencies, taking an average of $14.7 s$ to propose grasps in our experiments (see Section~\ref{subsec:results}).

Another family of approaches reconstructs the grasping surface into meshes~\cite{varley2017shapecompletion, lundell2020scenecompletion}, point clouds~\cite{chen2021transsc} or signed distance functions~\cite{breyer2020vgn, jiang2021synergies, cai2022volumetric} and uses the new representation to predict \mbox{6-DoF} grasps. One of these approaches, the Volumetric Grasping Network (VGN)~\cite{breyer2020vgn}, exhibits promising results by predicting grasp proposals from a truncated signed distance function (TSDF). Once the TSDF is built, grasp proposals can be predicted in real-time within $10 ms$, opening up possibilities for closed-loop control. 

VGN achieves a grasp success of $80\%$ using household objects in the original paper. However, the acquisition of the TSDF is achieved by integrating a stream of depth images acquired by a wrist-mounted depth camera, where the camera follows a pre-defined scan trajectory around a defined workspace prior to computing the grasp. When applying VGN to scenarios with an unknown camera-workspace transform, the approximate position of the object has to be estimated in advance to define the workspace around the object. Furthermore, if no wrist-mounted camera is available, the TSDF must be built from the incoming depth images of the head-mounted camera, which affects the quality of the TSDF. 

Another recent method, Contact-GraspNet~\cite{sundermeyer2021contactgraspnet}, proposes \mbox{6-DoF} grasps from a single point cloud. The method achieves a grasp success of more than $90\%$ in real-world experiments using a set of household objects. However, the model implementation requires a GPU with $\geq 8 GB$ RAM for running inference. Such requirements make Contact-GraspNet unsuitable for usage on mobile robotic platforms that do not have access to high-end dedicated GPUs, as is the case for several currently available mobile manipulators~\cite{pal2021tiago,yamamoto2018toyotahsr}.

From the literature, it is clear that few methods can be used directly to identify \mbox{6-DoF} grasps for unknown objects in unknown poses. In particular, no methods can be run on mobile manipulators without the need to run pre-processing, such as plane segmentation to filter grasp proposals or 3D object detection to identify the workspace, which can potentially lead to erroneous grasp proposals. We propose \ours\ to address these issues and provide a {ROS} package that can be used to generate \mbox{6-DoF} grasp proposals on mobile manipulators without any additional pre-processing.

\section{Grasp Proposal Network}\label{sec:network}

We consider the problem of proposing \mbox{6-DoF} grasps for a parallel-jaw gripper and using an {RGB-D} camera. The environment consists of a single object $o \in \mathcal{O}$ placed in a stable resting pose on a planar surface. The camera rests in an unknown, versatile pose near the object facing the planar surface. The goal is to propose a diverse set of \mbox{6-DoF} grasps $g \in \mathcal{G}$ for the object based on a depth image $\mathcal{I}$.

Similar to Sundermeyer et al.~\cite{sundermeyer2021contactgraspnet}, we represent grasps based on the contact point between the gripper and the object. Each pixel $(u, v)$ in a depth image $\mathcal{I}$ describes a potential grasp contact, i.e. the contact of a gripper plate during grasp execution. The full ground-truth grasp is defined as $g \in (u, v, q, \mathbf{r}, w)$, with the grasp quality $q$, the orientation of the grasp in camera coordinates $\mathbf{r}$ and the width of a grasp $w$. A visualisation of the grasp representation is depicted in Figure~\ref{fig:grasp} (b).

\begin{figure}[t]
      \centering
      \includegraphics[width=0.47\textwidth]{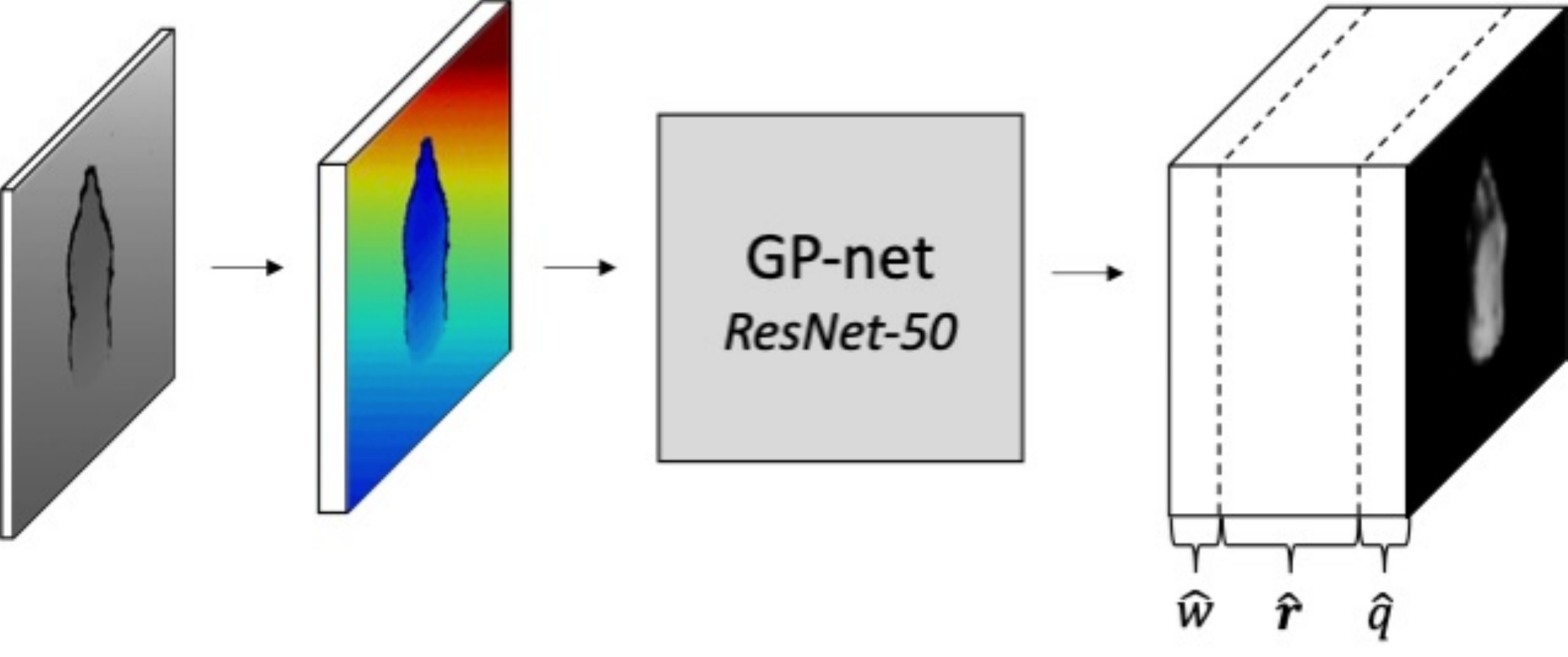}      
      \caption{Using a depth image as input, we apply a jet-colourscale, process the image with \ours\ and output pixel-wise grasp proposal predictions with a quality $\hat{q}$, orientation $\hat{\mathbf{r}}$ and grasp width $\hat{w}$.}
      \label{fig:network}
\end{figure}

We base the network architecture for \ours\ on a ResNet-50 model pre-trained on ImageNet. A description of the pipeline and the output tensor of the model can be seen in Figure~\ref{fig:network}. Since the rendered depth images consist of one channel and the pre-trained ResNet-50 architecture uses 3 input channels, we apply a jet-colourscale to the depth image as described by Eiter et al.~\cite{eitel2015rgbdobjectrecognition}. We set fixed normalisation boundaries of $0.4 m$ and $1.4 m$ to keep the relation to the real distance of a given scene. Note that points further away than $1.4 m$ are outside the grasping range of the robot.

\ours\ outputs a 6-channel tensor with the same spatial resolution as the input image, $W \times H$. Each pixel in the output tensor represents a grasp whose visible contact point corresponds to that pixel, with the channels containing predictions of the width of the grasp $\hat{w}$, the grasp orientation $\hat{\mathbf{r}}$ in form of a quaternion, and the quality of the grasp $\hat{q}$. Similar to VGN~\cite{breyer2020vgn}, we normalise the quaternions to unit quaternions and apply a sigmoid function to the quality channel.

\textbf{Loss function:} Since creating ground-truth grasp proposals for training at each pixel is computationally infeasible, we generate sparse maps containing grasp information at visible grasp contacts for up to $100$ pre-sampled grasps per object. The training data generation process is described in Section~\ref{sec:dataset}. We backpropagate the loss only through the output pixels at which we have ground-truth grasp information, similar to VGN~\cite{breyer2020vgn}. We define our loss function as:
\begin{align} \label{eq:loss}
    \mathcal{L} = \mathcal{L}_q + \mathbb{1}^{PosGrasp} (\alpha \mathcal{L}_\mathbf{r} + \beta \mathcal{L}_w)
\end{align}
with $\mathcal{L}_q$ being the binary cross-entropy loss between the ground truth and predicted binary quality values, $q$ and $\hat{q}$, $\mathcal{L}_\mathbf{r} = 1 - |\mathbf{r} \cdot \hat{\mathbf{r}}|$ the distance metric between the target and predicted quaternions, $\mathbf{r}$ and $\hat{\mathbf{r}}$, as defined in~\cite{kuffner2004metrics} and $\mathcal{L}_w$ the L1 loss between the ground-truth and predicted grasp width, $w$ and $\hat{w}$. Since unsuccessful grasps do not have valid configurations for the model to learn, we include the second part of the loss function only for ground-truth positive grasps indicated as $\mathbb{1}^{PosGrasp}$. We set $\alpha, \beta = 0.1$, which we experimentally found to give a good balance between $\mathcal{L}_q$, $\mathcal{L}_\mathbf{r}$ and $\mathcal{L}_w$.

In contrast to the loss function of VGN~\cite{breyer2020vgn}, we only allow one valid configuration for the orientation loss $\mathcal{L}_\mathbf{r}$. This change is rooted in the grasp contact representation, which anchors grasps to the visible grasp contact. Since the positive x-axis of the grasp coordinate frame points towards the second grasp contact (see Figure~\ref{fig:grasp}b), there remains only one valid orientation. This representation does not affect grasp variability since the parallel-jaw gripper is symmetric.

\textbf{Using the model output for robotic grasping:} To use the output of \ours\ for grasping objects with a robot, it must be transformed into grasp proposals. As a first step, a maximum of $j = 10$ grasps are chosen from the output tensor using non-maximum suppression with a peak distance of $4$ pixels and an acceptance threshold of $\gpthreshold = 0.4$. Note that the acceptance threshold $\gpthreshold$ for applying non-maximum suppression defines the minimum predicted quality $\hat{q}$ that will result in a grasp proposal and is chosen based on our ablation studies in Section~\ref{subsec:ablation}. Once the image coordinates $(u, v)$ for those grasp contacts are chosen, we re-project them into 3D camera coordinates using the camera intrinsics $K$ and the depth value of $(u, v)$ in the input depth image. Using the predicted quaternion $\hat{\mathbf{r}}$ and the predicted width $\hat{w}$ of each grasp, we translate the grasp contact to the tool centre point by moving them $0.5 \hat{w}$ along the grasp x-axis between the gripper plates. As a last step, the grasp is transformed from the camera coordinate frame to the robot base coordinate frame.

Our ROS package provides this functionality by running inference with a trained model, selecting the best grasps and mapping them to the robot base transform. Further, we provide example code to use the package with a pre-trained \ours\ model to plan grasps for a {PAL} {TIAGo} robot using a parallel jaw gripper. Using other gripper configurations, our code can generate an adjusted dataset, train a new model, and use it with our ROS package to produce grasp proposals.

\section{Training dataset}\label{sec:dataset}

To train \ours, we render a synthetic depth-image-based dataset with sparse ground-truth grasp information. Similar to DexNet2.0~\cite{mahler2017dexnet}, we use the objects from the 3Dnet \cite{wohlkinger20123dnet} and KIT \cite{kasper2012kit} mesh datasets. When loading the 3D meshes from the mesh datasets, we re-scale them to fit into TIAGo's gripper, a parallel gripper manufactured by PAL robotics with an opening width of $8 cm$~\cite{pal2021tiago}. Re-scaling 3D meshes is commonly used to generate robotic grasping datasets. In contrast to the re-scaling methods in \cite{mahler2017dexnet, morrison2020egad, eppner2021acronym}, we re-scale objects to a randomly chosen width, drawn uniformly from the range $6 cm$ to $10 cm$. In this way, our dataset includes objects which do not fit in our robotic gripper, which is a situation that will occur in real-world environments.

After re-scaling each object $o$, we calculate up to $25$ stable resting poses $\mathcal{S}(o)$ and use the antipodal grasp sampler proposed in~\cite{mahler2017dexnet} to generate up to $100$ parallel-jaw grasps $\mathcal{G}(o)$. Grasps are sampled by randomly choosing a surface point from the mesh and then sampling the grasp x-axis (see Figure~\ref{fig:grasp}b) based on the friction cone of the surface point. This procedure does not necessarily yield the best grasp for a grasp contact point. Our dataset should contain the best available grasps for a contact since we use it to train a network for proposing good grasps. Therefore, we modify the sampling procedure to generate $k = 6$ potential grasps for a given contact. We calculate the robust force closure metric $\robustness$~\cite{seita2016grasp, mahler2016dex, weisz2012grasp} for each of the sampled grasps at one contact point and keep the grasp with the highest $\robustness$. Using this method, we generate a total of $148,706$ ground-truth grasps for the object meshes in our dataset.

Since the robust force closure metric $\robustness$ of a sampled ground-truth grasp depends on the grasp contacts, it is defined by the grasp x-axis and consistent for any grasp approach axis, i.e. the direction of the z-axis in Figure~\ref{fig:grasp}. However, collisions between the gripper, object and planar surface further influence the success of a grasp. We define the quality $q$ of a grasp in our rendered dataset as
\begin{align} \label{eq:quality}
    q(g) = \begin{dcases} 1 \quad \robustness \geq \threshold \text{ and } coll\_free(g) \\ 0 \quad otherwise \end{dcases}
\end{align}
with  $\threshold = 0.5$ being the robustness threshold and $coll\_free(g)$ indicating if a grasp is collision free, similar to DexNet2.0~\cite{mahler2017dexnet}.

We aim to choose reproducible grasp orientations for a given scene and ground-truth grasp. To achieve this, we apply the following steps for selecting the grasp approach axis orientation for a given grasp: To check for collisions, we hinge-rotate the grasps in steps of $\Delta\grasphinge = 15\deg$ around the contact points and thereby the grasp x-axis (see Figure~\ref{fig:grasp}b). If we have collision-free grasps within those orientations, we set the ground-truth grasp orientation as the median collision-free grasp. If we have multiple collision-free regions, we choose the median collision-free grasp whose approach is most closely aligned with the principal ray of the camera, e.g. approaching from the front of the object rather than from behind the object. With these steps, we fully define each grasp orientation in our dataset in a reproducible way.

We render $n = 20$ images for each stable pose $s \in \mathcal{S}(o)$ of each object $o \in \mathcal{O}$ from camera poses selected uniformly at random as described in VGQ-CNN~\cite{konrad2022vgqcnn}. We then project the grasp contacts into the image plane and calculate the image coordinates of the visible contact $(u, v)$. If multiple collision-free grasp contacts are visible at one pixel, we choose the grasp with the highest robust force closure value $\robustness$. In addition, we store the binary segmentation mask of the object and use the pixels not showing the object as ground-truth negative grasp contacts during training. The full dataset consists of $260,340$ images with an average of $88.1$ grasps per image, totalling $22,944,376$ grasps.

\section{Experiments}

We train \ours\ for 20 epochs using an Adam optimiser with a learning rate of $3e^{-4}$ and a batch size of $32$. %Using two NVIDIA GeForce RTX 3090 GPUs, the training takes approximately $1.5$ days to complete. 
To reduce the sim-to-real-gap for \ours, we simulate depth-camera noise on our depth images using the noise model described in~\cite{handa2014kinectnoise, Barron2013kinectnoise}. We find that compared to the Gaussian noise model suggested in DexNet2.0~\cite{mahler2017dexnet}, the added depth-camera noise model substantially improves the robustness of \ours\ in real-world scenarios.

\subsection{Results}\label{subsec:results}

We evaluate the performance of \ours\ with simulation and real-world experiments using objects from the EGAD evaluation benchmark~\cite{morrison2020egad} not seen by the network during training. The simulation analysis is used to validate our model and conduct the ablation studies in Section~\ref{subsec:ablation} since it requires less time and fewer resources than tests with a real robot. For the real-world experiments, we use \ours\ on a {PAL} {TIAGo} mobile manipulator to grasp the $49$ {3D} printed objects from the EGAD evaluation benchmark~\cite{morrison2020egad}. We repeat the real-world experiments with VGN~\cite{breyer2020vgn} and GPD~\cite{pas2018detect} and compare them to the performance of \ours.

\begin{figure}[t]
      \centering
      \includegraphics[width=0.3\textwidth]{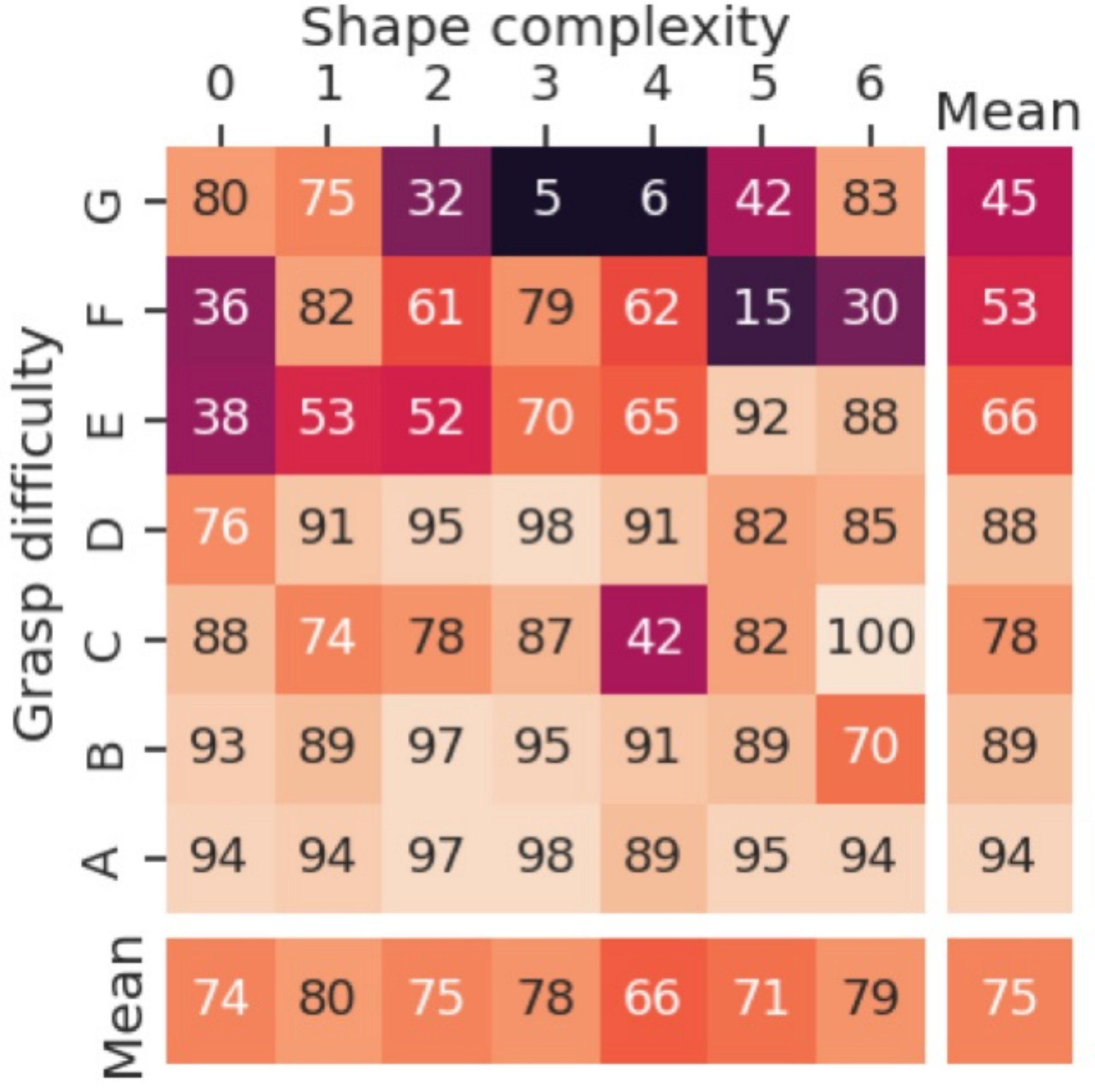}
      \caption{Grasp success [\%] in simulation using \ours\ to grasp objects from EGAD with a {PAL} parallel jaw gripper.}
      \label{fig:sim_results}
\end{figure}

\begin{figure*}[t]
    \centerline{\includegraphics[width=0.8\textwidth]{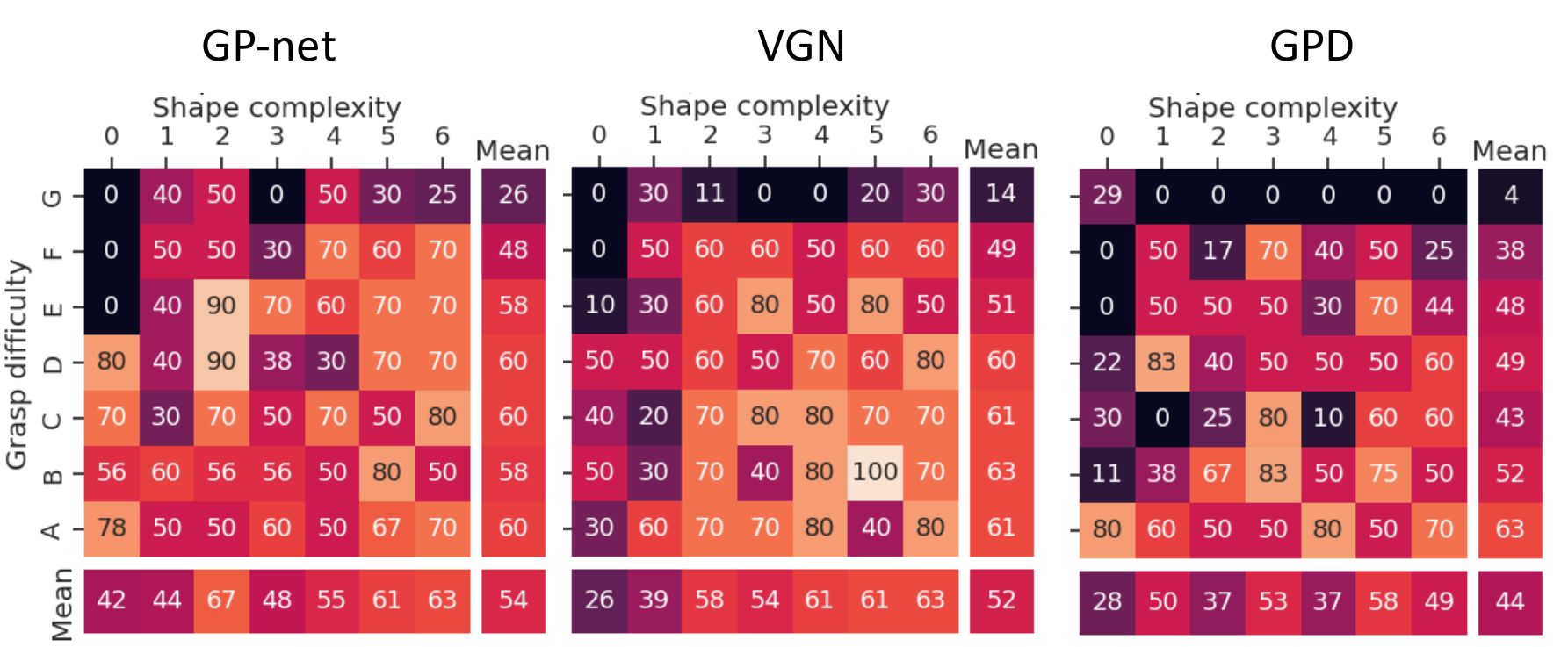}}
      \caption{Grasp success [\%] out of 10 grasp attempts per object on the EGAD benchmark using GP-net, VGN and GPD for grasp proposal.}
      \label{fig:res}
\end{figure*}

\textbf{Simulation experiments:} We test the grasp success of the grasps proposed by \ours\ in simulation using the pybullet~\cite{coumans2021pybullet} physics engine based on the simulation environment developed for VGN~\cite{breyer2020vgn}. A parallel jaw gripper is simulated to approach the grasp pose linearly along the grasp z-axis for $0.15 m$, close the gripper using a maximum force of $5 N$, and lift the object. The grasp is labelled as successful if the gripper can lift the object $0.1 m$ above the planar surface. 
For each trial, one of the $49$ objects in the EGAD evaluation benchmark is placed in a random position within the workspace within a $30\times30 cm^2$ workspace on a planar surface. Then, a depth image is rendered from a camera pose sampled uniformly at random from the spherical coordinates around the centre of the workspace with the sampling bounds used for training in Section~\ref{sec:dataset}. We run inference with \ours\ on the depth image with added depth-camera noise and map the output tensors to grasp proposals as described in Section~\ref{sec:network}. 

We simulate the grasp with the highest predicted grasp quality $\hat{q}$ proposed by the model for one trial. Each experimental run comprises $100$ such trials for each object from the EGAD~\cite{morrison2020egad} evaluation set. In the simulation, \ours\ achieves a mean grasp success of $74.6\%$ across all objects of the EGAD dataset. The grasp success on each object is depicted in Figure~\ref{fig:sim_results}. The shape complexity and grasp difficulty of each object are defined in the EGAD evaluation set. Each object is named according to its shape complexity and grasp difficulty, e.g. object ``A0'' with grasp difficulty ``A'' and shape complexity ``0''. The grasp success decreases with increasing grasp difficulty, while it is consistent across the levels of shape complexity. 

\textbf{Real-world experiments:} We test the grasp success in real-world experiments with a {PAL} TIAGo mobile manipulator~\cite{pal2021tiago}. We use an Intel Realsense D435 depth camera mounted on TIAGo's head since it has a lower minimum depth distance, and we find it to give a better performance on thin features than the Orbbec Astra used in the TIAGo mobile manipulator by default. We average over ten consecutive depth image frames for noise reduction for all tested methods. 

We use the same $49$ evaluation objects from the EGAD dataset~\cite{morrison2020egad} as in our simulation analysis for the experiments. For each object, a total of 10 grasp trials are executed, resulting in $490$ grasp attempts for each algorithm. We use two different initial robot poses for grasp-planning, each having a different head tilt and torso height and, thereby, a different camera viewpoint of the scene. The object is dropped within a $30 \times 30 cm^2$ square on the table by a human operator before each grasp trial.

We choose VGN~\cite{breyer2020vgn} and GPD~\cite{pas2018detect} to compare to \ours\ in real-world experiments, given their use as points of comparison by other researchers~\cite{jiang2021synergies, cai2022volumetric}. To apply VGN to our experiments, we define the pose of the $30\times30\times30 cm^3$ workspace for grasp sampling in front of the robot manually to sit on the table. Note that VGN was originally intended to be used with a wrist-mounted depth camera performing a grasp scan along a trajectory to build the TSDF. Since our work focuses on proposing grasps from a single viewpoint, we instead apply VGN by building the TSDF from a single, noise-reduced depth image. 

To apply GPD to our experiments, we re-project a point cloud from a single, noise-reduced depth image and crop it to reduce the number of points and improve run-time. Further, the object points in the point cloud are indexed to indicate where grasps should be sampled. The indexing of the point cloud is achieved by fitting a plane to the point cloud and indexing points with a distance $\geq 0.005 m$ to the table plane. While the approaches for setting the workspace for VGN and indexing the point cloud for GPD work for our experimental setting, more general applications would require more sophisticated workarounds to prevent erroneous grasp proposals. \ours\ does not need any workspace definition and can be directly applied to the depth image.

\begin{table}[t]
    \centering
    \begin{tabular}{c|c c c}
                     &  GP-net & VGN~\cite{breyer2020vgn} & GPD~\cite{pas2018detect}\\
                     \hline
        Grasp success [\%] & $\mathbf{54.4}$ & $51.6$ & $44.2$ \\
        Inference time [s] & $2.1$ & $\mathbf{1.2}$ & $14.7$ \\
        Grasp planning time [s] & $\mathbf{2.7}$ & $4.9$ & $19.1$ \\
    \end{tabular}
    \vspace{0.3cm}
    \caption{Grasp success, inference, and grasp planning time for our real-world experiments. Inference refers to the time between network input and output, while grasp planning additionally includes pre-processing and post-processing.}
    \label{tab:results}

    \vspace{-0.6cm}
\end{table}

We run the algorithms for all methods on an Intel i7-10750H CPU and NVIDIA RTX 2060 GPU. The results are shown in Figure~\ref{fig:res}. \ours\ performs slightly better than VGN and substantially better than GPD with a grasp success of $54.4\%$ compared to their $51.6\%$ and $44.2\%$, respectively. We report each network's overall results and timings in Table~\ref{tab:results}. In our experiments, \ours\ takes $2.1$s for proposing grasps on a given depth image, while VGN and GPD require $1.2$s and $14.7$s, respectively. These times do not include additional pre-processing steps like table segmentation and workspace definition, which are necessary for VGN and GPD. Due to this, for the overall grasp planning time, which includes pre-processing, post-processing and sending data between different {ROS} nodes, \ours\ is faster than VGN and GPD with $2.7s$ compared to $4.9s$ and $19.1s$, respectively. 

Note that experimental results are sensitive to the hardware and software used and thereby make it difficult to compare methods if not tested in the same setting. Further, real-world experiments are expected to result in a lower grasp success than our simulation results, which are simulated only with the end-effector. These performance differences are rooted in perception and actuation uncertainties, collision checking, and the performance of the path planning algorithms.

\begin{figure}[t]
      \centering
      \includegraphics[width=0.47\textwidth]{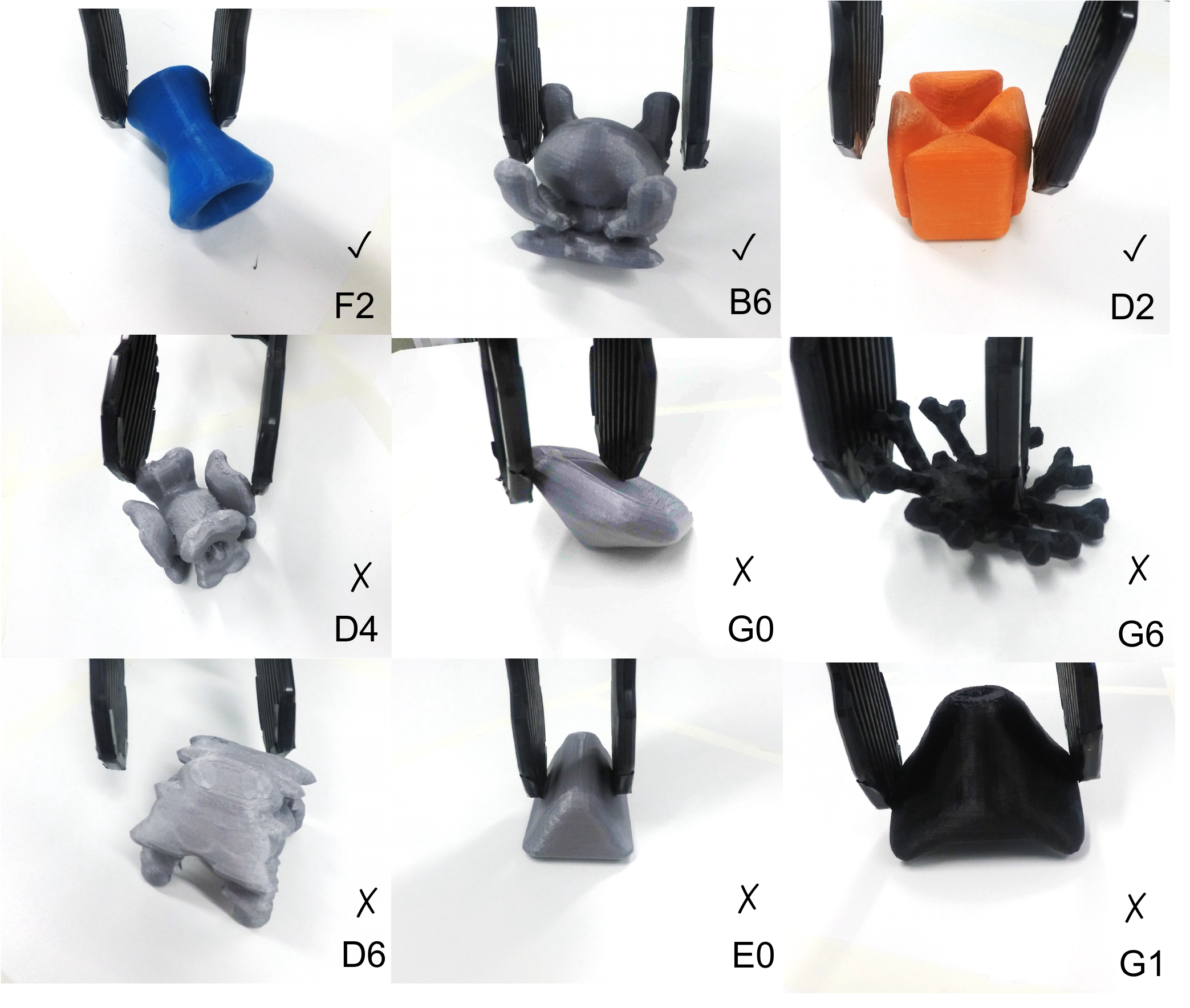}
      \caption{Success and failure cases of using \ours\ to propose grasps on objects of the EGAD evaluation set with a {PAL} {TIAGo} mobile manipulator. We label each figure according to the corresponding object in the dataset and indicate success with \checkmark and failures with X.}
      \label{fig:failure}
\end{figure}

We have encountered different failure cases when testing \ours\ in our real-world experiments. An overview of different successful grasp attempts and failures is shown in Figure~\ref{fig:failure}. One type of failure occurs when the gripper can not close fully due to obstructions by itself or the object, e.g. in the examples of objects ``G0'' and ``G6''. The object will lose contact with the gripper once lifted, and the grasp attempt will fail. 

In other failure cases, the gripper pushes the object away when approaching due to perception and actuation uncertainties, as seen in the example of objects ``D4'' and ``D6''. This type of failure is hardware-dependent and not necessarily rooted in the grasp proposal algorithm. Further, it could be prevented when using closed-loop control, e.g., using an RGB camera mounted on the gripper and correcting the grasp approach. The grasps shown with objects ``E0'' and ``G1'' fail because the friction between the objects and the gripper is too low to compensate for the slanted surfaces that are being grasped. Note that grasping ``E0'' longitudinally is not an option, as this side of the object is longer than the gripper width, and therefore, ``E0'' fails to be grasped in most cases.

\subsection{Ablation studies}\label{subsec:ablation}

We conduct a set of ablation studies to investigate the performance of \ours further. We investigate how an alternative grasp representation defining the Tool Centre Point (TCP) performs compared to the contact-based grasp representation (see Figure~\ref{fig:grasp}b) used for \ours. Further, we run simulation experiments and investigate how different acceptance thresholds $\gpthreshold$ for the non-maximum suppression influence the grasp success and the final number of grasp proposals.

\textbf{Grasp representation:} The grasp representation used in \ours\ is contact-based, with each pixel representing a visible grasp contact of a potential grasp, see Section~\ref{sec:network}. This grasp representation was first proposed in Contact-Graspnet~\cite{sundermeyer2021contactgraspnet}, suggesting it facilitates the learning process by reducing dimensionality. However, the grasp representation in~\cite{sundermeyer2021contactgraspnet} was not compared to conventional, TCP-based grasp representations as used in most of the related work~\cite{breyer2020vgn, pas2017pointclouds, lundell2020scenecompletion, mahler2017dexnet,satish2019policy, morrison2019, redmon2015, kumra2020, zhang2020randomforest, berscheid20216dof}. 

While the contact-based grasp representation can potentially increase the model performance, using this grasp representation can lead to problems where no grasp contact is visible. For example, if a robot is supposed to grasp a book but is facing the spine of the book head-on. In this situation, the areas for potential grasp contacts on the left and right sides are not visible, and a contact-based method would not be able to produce valid grasp configurations. Here, a repositioning of the camera would be necessary.

For this reason, we compare the performance of a contact-based grasp representation to a TCP-based grasp representation. We modify \ours\ to use a TCP-based grasp representation as $g \in (u, v, z, q, \mathbf{r}, w)$ with $(u, v)$ being the image coordinates of the grasp centre and $z$ being the distance between the grasp centre and the visible surface depth at $(u, v)$. We train the model on our dataset as described in Section~\ref{subsec:results} and compare performance to \ours\ using the simulation analysis. Due to the extra variable $z$ in the grasp representation, the model has seven output channels, and we define the adapted loss function as:
\begin{align}
    \mathcal{L} = \mathcal{L}_q + \mathbb{1}^{PosGrasp} (\alpha \mathcal{L}_\mathbf{r} + \beta \mathcal{L}_w + \nu \mathcal{L}_z)
\end{align}
with $\mathcal{L}_z$ being the L1 loss between ground-truth $z$ and predicted grasp distance $\hat{z}$. We set $\alpha, \nu = 0.1$ and $\beta = 0.01$ since the width is not necessary to predict the grasp position for the TCP-based grasp representation. After training the model for $20$ epochs, we achieve a mean grasp success of $27.2\%$ in simulation, substantially lower than \ours's $74.6\%$ with the contact-based grasp representation. We find that the orientation loss $\mathcal{L}_\mathbf{r}$ plateaus almost double that of \ours's loss with $0.24$ for the TCP-based grasp representation and $0.13$ with the contact-based grasp representation.

We hypothesise that this difference is rooted in the reduction of ambiguities when grasps are anchored to their grasp contact points. Ground-truth positive grasps would have similar orientations for neighbouring grasp contacts on an object, while TCP-based grasps could approach the same TCP position from various angles. This is especially apparent when looking at cylindrical objects, where a TCP-based ground truth positive grasp could rotate the grasp axis to any orientation around the centre of the cylinder. In contrast, a contact-based ground truth positive grasp only permits one orientation such that the x-axis of the grasp passes through the centre of the cylinder (see Figure~\ref{fig:grasp}b).

\begin{figure}[t]
      \centering
      \includegraphics[width=0.47\textwidth]{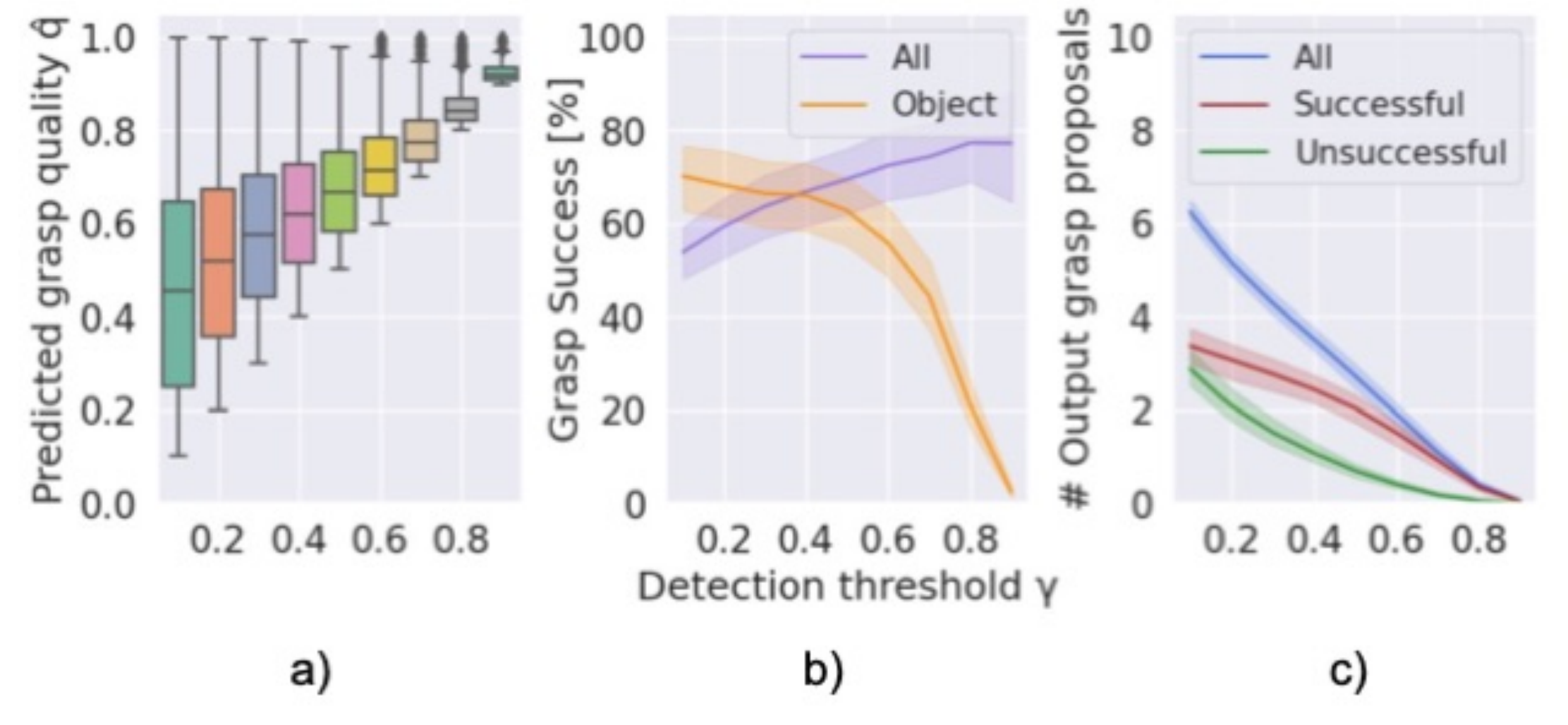}
      \caption{Simulation results showing the influence of varying the acceptance threshold $\gpthreshold$ on a) the predicted quality $\hat{q}$ of grasp proposals, b) the grasp success and c) the number of output grasp proposals per object with a $95\%$ confidence interval.}
      \label{fig:abl_threshold}
\end{figure}

\textbf{Acceptance threshold $\gpthreshold$:} The acceptance threshold $\gpthreshold$ defines the minimum predicted grasp quality $\hat{q}$ of output grasp proposals for \ours. Non-maximum suppression with $\gpthreshold$ as a cut-off point applied to the predicted grasp quality $\hat{q}$ yields the output grasp proposals. As such, $\gpthreshold$ balances the number of proposed grasps and the confidence the model has in those grasps. Ideally, $\gpthreshold$ should be set to propose as many grasps as possible while retaining good confidence in the quality of those grasps. To investigate how this trade-off affects \ours, we apply a range of acceptance thresholds $\gpthreshold = [0.1, 0.2, \dots, 0.9]$ and investigate the influence on grasp performance in simulation.

The results can be seen in Figure~\ref{fig:abl_threshold}. Increasing $\gpthreshold$ increases the predicted grasp quality $\hat{q}$ in Figure~\ref{fig:abl_threshold}a) and decreases the number of output grasp proposals in Figure~\ref{fig:abl_threshold}c). The number of unsuccessful grasp proposals decreases non-linearly when increasing $\gpthreshold$. We show both the grasp success of all proposed grasps and the grasp success for each object in Figure~\ref{fig:abl_threshold}b). Note that the grasp success for each object uses only the output grasp proposal with the highest predicted quality $\hat{q}$ and reports an unsuccessful grasp if there is no grasp proposal, as explained in Section~\ref{subsec:results}. The two curves show the trade-off between not predicting any grasp and thereby failing to pick up the object, i.e. a low grasp success for the object, and the success of all grasps proposals, i.e. a lower grasp success on all grasp proposals if many unsuccessful grasps are proposed. 

To balance the number of grasps proposed with the success of those grasps, we set $\gpthreshold = 0.4$ for all of our other experiments.

\section{Discussion and Conclusion}

In this paper, we presented \ours, a model to propose \mbox{6-DoF} grasps based on depth images from flexible and unknown viewpoints. In contrast to widely used algorithms for robotic grasping like GPD~\cite{pas2017pointclouds} and VGN~\cite{breyer2020vgn}, the viewpoint flexibility enables \ours\ to be used without the need to pre-define a workspace or filter grasps to produce good grasp proposals. Hence, \ours\ can be used on mobile robots without additional pre-processing steps like plane-fitting, table segmentation or 3D object detection. To train \ours, we created a synthetic dataset based on more than $1400$ object meshes with ground-truth grasp information. 

We evaluated \ours\ using a {PAL} {TIAGo} mobile manipulator in real-world experiments and compared its performance to GPD and VGN on the EGAD evaluation benchmark~\cite{morrison2020egad}. \ours\ achieved a grasp success of $54.4\%$, slightly better than VGN's $51.6\%$ and substantially higher than GPD's $44.2\%$ in our setup while not requiring any of the pre-processing steps for workspace definition and grasp filtering necessary for VGN and GPD.

The contact-based grasp representation utilised in \ours\ can not represent grasps without a visible grasp contact. This limitation is not existent in TCP-based grasp representations. To weigh the benefits of a contact-based grasp representation against these limitations, we compare it to a TCP-based grasp representation in our ablation studies in Section~\ref{subsec:ablation}. We find a significant performance advantage of a model trained with the contact-based representation.

When using a mobile manipulator in real-world scenarios, the robot is able to move the camera and view objects from different perspectives. In this process, previously invisible grasp contacts can be made visible. Under these circumstances, we think the benefits of a contact-based representation outweigh its limitations.

Similar to most of the commonly used methods~\cite{sundermeyer2021contactgraspnet, breyer2020vgn, lundell2020scenecompletion}, a limitation of \ours\ is the gripper-dependence of our dataset and hence the model. Since the quality of grasps depends on the grasp being collision-free, the design of the gripper is used implicitly during dataset generation when checking for collisions, see Eq.~\ref{eq:quality}. The resulting model learns the implicit gripper design, and hence, the performance with alternative gripper configurations cannot be guaranteed.

VGN and GPD work in cluttered environments, while GP-net is trained for single objects. In this work, we proved the method of our approach to propose \mbox{6-DoF} grasps from versatile, unknown viewpoints without workspace definition. In future work, we plan to extend GP-net to cluttered scenes, which has been shown to work for image-based grasp quality prediction algorithms~\cite{mahler2017clutter}. Further, we aim to base this extension of \ours\ to be used in more diverse and realistic scenarios with various furniture units, e.g. desks, sideboards or shelves.

\bibliographystyle{IEEEtran}
\bibliography{references}

\end{document}

%% file: nomenclature.tex
\newcommand{\robustness}{\epsilon}

\newcommand{\grasphinge}{\omega}

\newcommand{\threshold}{\delta}
\newcommand{\gpthreshold}{\gamma}